%% file: arxiv.tex
\documentclass[10pt,twocolumn,letterpaper]{article}

\usepackage[pagenumbers]{cvpr} %

\input{preamble}

\definecolor{cvprblue}{rgb}{0.21,0.49,0.74}
\usepackage[pagebackref,breaklinks,colorlinks,allcolors=cvprblue]{hyperref}

\def\confName{CVPR}
\def\confYear{2027}

\newcommand{\datasetlink}{https://huggingface.co/datasets/clinicalspace/Clinical3D-Bench}

\title{\bench: Evaluating and Improving Spatial Intelligence of\\Multimodal LLMs in Hospital Environments}

\author{Yang Bai\textsuperscript{1*}\quad
Yingzhi Xia\textsuperscript{1*}\quad
Ling Lin\textsuperscript{2}\quad
Jun Cheng\textsuperscript{1}\quad
Lin Fu\textsuperscript{3}\quad
Huazhu Fu\textsuperscript{1\dag}\\
\textsuperscript{1}IAIC, A*STAR\qquad
\textsuperscript{2}University of Science and Technology of China\qquad
\textsuperscript{3}Wenzhou Medical University\\
{\normalsize Dataset: \url{\datasetlink}}}

\begin{document}
\input{sec/teaser}
{\let\thefootnote\relax\footnotetext{\textsuperscript{*}Equal contribution.\quad \textsuperscript{\dag}Corresponding author.\quad \textsuperscript{3}National Clinical Research Center for Ocular Diseases, Eye Hospital, Wenzhou Medical University, Wenzhou, 325027, China.}}
\input{sec/0_abstract}
\input{sec/1_intro}

\input{sec/2_related}
\input{sec/3_benchmark}
\input{sec/4_evaluation}
\input{sec/5_geojev}
\input{sec/6_conclusion}
{
    \small
    \bibliographystyle{ieeenat_fullname}
    \bibliography{main}
}

\clearpage
\appendix
\twocolumn[\centering\Large\bfseries \bench: Supplementary Material\vspace{1.5em}]
\input{sec/X_suppl}

\end{document}

%% file: preamble.tex
\usepackage[T1]{fontenc}
\usepackage{cuted}      %
\usepackage{caption}    %
\usepackage{multirow}
\usepackage{makecell}
\usepackage{pifont}     %
\usepackage{bm}
\usepackage{array}
\usepackage{colortbl}
\newcolumntype{L}[1]{>{\raggedright\arraybackslash}p{#1}}

\newcommand{\bench}{Clinical3D-Bench\xspace}
\newcommand{\trainset}{Clinical3D-45K\xspace}
\newcommand{\geojev}{GeoJeV\xspace}
\newcommand{\videojev}{VideoJev\xspace}

\newcommand{\cmark}{\ding{51}}
\newcommand{\xmark}{--}
\definecolor{rowblue}{RGB}{225,235,245}    %
\newcommand{\grp}[1]{\rowcolor{rowblue}\textit{\textbf{#1}} & & & & & & & & & & \\}
\newcommand{\fr}[1]{\,{\scriptsize\color{gray}(#1\,f)}}   %

\DeclareMathOperator*{\argmax}{arg\,max}
\DeclareMathOperator{\softmax}{softmax}
\newcommand{\pM}{p_{\mathrm{M}}}
\newcommand{\sg}{\mathrm{sg}}

%% file: sec/teaser.tex
\twocolumn[{%
\renewcommand\twocolumn[1][]{#1}%
\maketitle
\vspace{-14pt}
\begin{center}
\includegraphics[width=\textwidth]{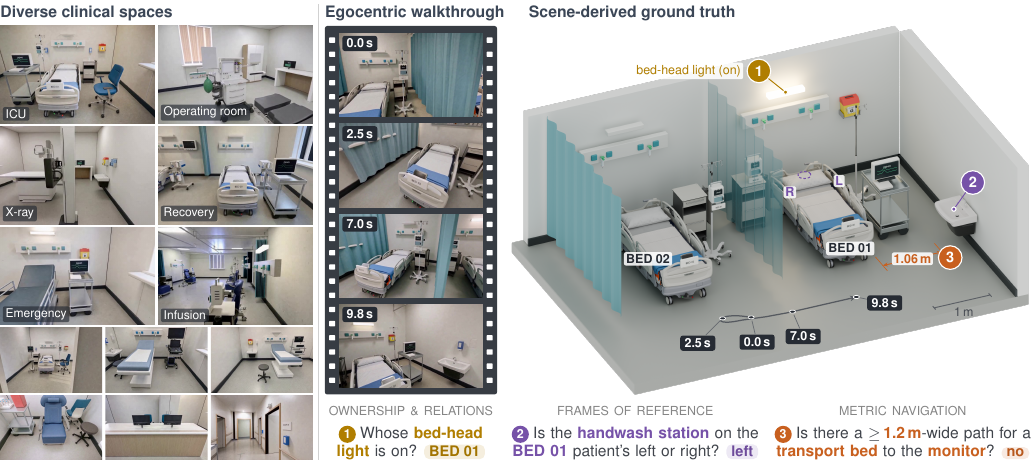}
\captionsetup{hypcap=false}
\captionof{figure}{\textbf{\bench.} An MLLM answers questions from a hospital walkthrough (middle), with exact answers computed from the 3D scene (right; numbers mark evidence). Three of eight question types illustrate clinical ownership, patient-centered reference frames, and equipment clearance (\cref{fig:examples}). Left: selected examples from the 16 room types.}
\captionsetup{hypcap=true}
\label{fig:teaser}
\end{center}
\vspace{4pt}
}]

%% file: sec/0_abstract.tex
\begin{abstract}
Spatial reasoning in hospitals requires tracking look-alike
equipment, interpreting patient-relative directions, and
judging passage widths, yet existing benchmarks provide
limited coverage of these abilities.
We introduce \textbf{\bench}, a simulated benchmark with 3,368
questions of eight types on egocentric walkthroughs of 373
hospital-room layouts.
Layouts follow hospital design guidelines, answers are
computed from the 3D scenes, and paired scene states test
whether models change their answers when relevant evidence
changes.
Videos are photorealistic, and ten physicians screened the
layouts.
Among 35 multimodal large language models (MLLMs), the best
open and proprietary models reach 45.7\% and 53.6\% macro
accuracy, against 40.8\% for an answer prior and 79.0\% for
physicians on a subset, and 27 of 32 open models score below
40\% on two-way patient-relative directions.
Fine-tuning on \textbf{\trainset}, our 44,750-question training set, lifts two MLLMs 16--20 points
above the best proprietary model.
3D maps given to the fine-tuned MLLM as input go largely unused: replacing them with another video's,
even when they are ground truth, barely changes its accuracy.
The proposed \textbf{\geojev} instead lets the MLLM compute with 3D: it reconstructs a scene memory
from each video, selects 3D operations and their arguments from the MLLM's question state, and adds
their evidence to its option scores.
It achieves the highest accuracy of all evaluated models, narrowing the gap to physicians.
\end{abstract}

%% file: sec/1_intro.tex
\section{Introduction}
\label{sec:intro}

Hospitals are 3D environments where small spatial errors carry clinical consequences~\cite{wrongsite,wrongside}.
A nurse walking through a ward knows which bed a lit bed-head light belongs to, what lies on the patient's left, and whether a transport bed can pass a bedside cart. MLLMs supporting hospital assistants need to make these judgments from egocentric video. This is challenging because similar equipment repeats across beds, the patient's left and right may differ from the viewer's, and passage depends on both obstacle geometry and equipment width.

Existing benchmarks evaluate spatial reasoning in general scenes~\cite{simsv,viewspatial,ostbench,sat,vsibench,mmsibench}, but provide limited coverage of patient-relative directions, equipment ownership, and width-constrained passage. This requires videos with clear visual evidence and accurate spatial annotations. Collecting such data in hospitals is difficult because of patient privacy and the cost of measuring scene geometry. We therefore use simulation to construct hospital scenes with controlled layouts and scene-derived answers.

We introduce \textbf{\bench} (\cref{fig:teaser}), with 3,368 questions across eight question types and 373 layouts spanning 16 hospital room types (\cref{fig:examples}). Layouts follow constraints derived from hospital design guidelines, and answers are computed from the underlying 3D scenes, including routes and passage widths. Paired states of the same layout differ in one scene condition, such as which bed-head light is on, allowing us to test whether models change their answers when the relevant evidence changes. Rendered walkthroughs are converted to photorealistic videos and checked for camera-path consistency. Ten physicians screened the layouts, reviewing one video per layout. We also provide \trainset, a training set with 44,750 questions from layouts disjoint from the benchmark.

We evaluate 32 open MLLMs and three proprietary models, and fine-tune two models on \trainset. 
The best open and proprietary models reach 45.7\% and 53.6\% macro accuracy. For reference, a test-fitted answer prior scores 40.8\%, and physicians score 79.0\% on a 400-question subset. Patient-relative directions are particularly challenging: 27 of the 32 open zero-shot models score below 40\%, against 50\% chance.
Fine-tuning VideoJev-4B~\cite{videojev} raises its macro accuracy from 45.5\% to 73.9\% and its patient-frame accuracy from 17.0\% to 93.7\%. On paired states with different correct answers, evaluated zero-shot models retain the same answer more often than their answer preferences predict; fine-tuning reduces this excess repetition. However, route planning and passability remain the fine-tuned model's weakest tasks, at 58.3\% and 66.0\%, respectively.

These results motivate testing whether explicit 3D input
\cite{vlm3r,gpt4scene,spatialmllm} can improve spatial reasoning. However, adding bird's-eye-view (BEV) tokens~\cite{liftsplat}, local map reads, or a geometry head during MLLM fine-tuning yields no clear gain in overall accuracy. For BEV tokens and local map reads, replacing the correct map with another clip's barely changes accuracy; the ground-truth map setting shows the same pattern. These findings motivate computing evidence from 3D representations and using it directly to score answer options.

To this end, we introduce \textbf{\geojev} (\cref{sec:geojev}, \cref{fig:method}), which reconstructs a scene memory from each video and uses three 3D primitives for object-track queries, free-space analysis, and learned BEV map reads. Linear heads on the fine-tuned VideoJev-4B's question state select a primitive and its arguments, and the resulting evidence is added to the model's option scores. This requires no second language model. \geojev improves macro accuracy from 73.9\% to 75.7\%, with gains mainly in counting, re-identification, passability, and direction; route accuracy improves only from 58.3\% to 59.5\%.
Replacing its scene memory with another layout's lowers accuracy to 70.4\%, showing that its predictions depend on scene-specific evidence.

Our contributions are threefold:
(1) \textbf{\bench}, a hospital spatial reasoning benchmark with scene-derived answers and paired scene states, accompanied by \trainset.
(2) An evaluation of 35 MLLMs and fine-tuning experiments that reveal persistent difficulties in route planning and passability, alongside limited gains from the tested 3D input strategies.
(3) \textbf{\geojev}, a framework that selects 3D computations from an MLLM's question state and combines their evidence with option scores, improving accuracy over the fine-tuned baseline.

%% file: sec/2_related.tex
\section{Related work}
\label{sec:related}

\paragraph{Spatial benchmarks for MLLMs.} Existing benchmarks test spatial reasoning from videos~\cite{vsibench}, multiple images~\cite{mmsibench}, online observations~\cite{ostbench}, and different
viewpoints~\cite{viewspatial}. VSI-Bench includes human-annotated routes. SAT and SIMS-V generate spatial questions through simulation~\cite{sat,simsv}, while Open3D-VQA uses real and simulated aerial scenes~\cite{open3dvqa}. Clinical3D-Bench evaluates patient-relative directions, equipment relations, and routes and passage widths computed from hospital scene occupancy. Paired scene states with different correct answers test whether models respond to changes in the scene (\cref{tab:compare}).

\paragraph{Clinical spatial reasoning.} Medical VQA benchmarks focus on medical images~\cite{vqarad,pathvqa,omnimedvqa}, including 3D spatial reasoning in SpatialMed~\cite{spatialmed}. MM-OR and EgoExOR study surgical scenes and activities~\cite{mmor,egoexor}, while OR-Expert addresses operating-room spatial reasoning \cite{orexpert}. Clinical3D-Bench complements these tasks with patient-relative directions, equipment ownership, and clearance-constrained navigation across diverse hospital rooms.

\begin{table*}[t]
\centering
\caption{\textbf{Comparison with related benchmarks.} \#QA is the total number of questions; for \bench it counts the test set, which comes with \trainset (44,750 training questions). In brackets: whose frame of reference the questions use.}
\label{tab:compare}
\setlength{\tabcolsep}{4pt}
\resizebox{\textwidth}{!}{%
\begin{tabular}{@{}llllrlccc@{}}
\toprule
Benchmark & Domain & Appearance & Input & \#QA & \makecell[l]{Exact geometric\\labels} & \makecell{Answer-changing\\states} & \makecell{Perspective\\taking} & \makecell{Routes /\\clearance} \\
\midrule
VSI-Bench~\cite{vsibench} & homes, offices & real scans & video & 5k & from 3D annotation & \xmark & \cmark{} (viewer) & route plan \\
MMSI-Bench~\cite{mmsibench} & mixed & real images & multi-image & 1k & human-written & \xmark & partly & \xmark \\
OST-Bench~\cite{ostbench} & indoor rooms & real scans & online video & 10k & from 3D annotation & \xmark & \xmark & \xmark \\
ViewSpatial-Bench~\cite{viewspatial} & mixed & real images & image & 5.7k & from annotation & \xmark & \cmark{} (person) & \xmark \\
SAT~\cite{sat} & homes & rendered & image(s) & 175k & \cmark & partly & \cmark{} (viewer) & \xmark \\
SIMS-V~\cite{simsv} & homes & rendered & video & 203k & \cmark & \xmark & \cmark{} (viewer) & \xmark \\
Open3D-VQA~\cite{open3dvqa} & cities (aerial) & real + rendered & image, point cloud & 73k & \cmark{} (rendered part) & \xmark & \xmark & \xmark \\
\midrule
\textbf{\bench} & \textbf{hospital rooms} & \textbf{photorealistic (from renders)} & \textbf{video} & 3.4k & \cmark & \cmark & \cmark{} (patient) & \cmark \\
\bottomrule
\end{tabular}}
\end{table*}

\begin{figure*}[t]
\centering
\includegraphics[width=\textwidth]{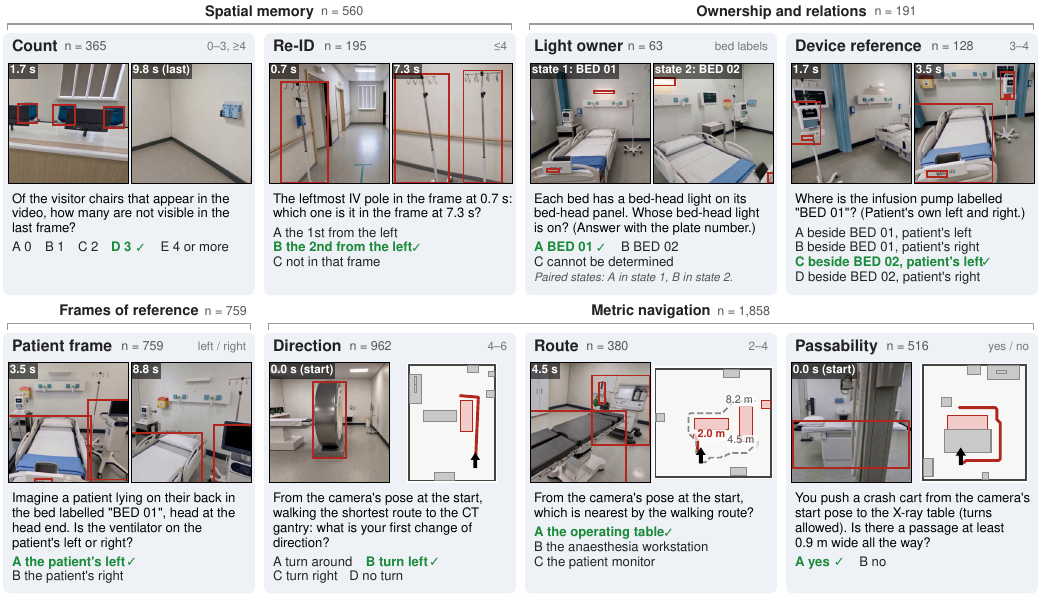}
\caption{\textbf{The eight question types of \bench}, organized into four task groups, with one example each (wording abridged).
$n$ is the number of questions; the top right of a card gives the answer options. Red boxes mark the ground-truth
evidence and green the answer. Navigation cards add a top view with the camera's start pose (arrow), the goal (red) and
the shortest walk.}
\label{fig:examples}
\end{figure*}

\paragraph{3D-aware MLLMs and tools.}
MLLMs incorporate geometry through reconstruction features~\cite{vlm3r,spatialmllm}, position encodings~\cite{video3dllm,llava3d}, BEV prompts~\cite{gpt4scene}, or spatial supervision~\cite{ross3d,spatialladder}. Other approaches use 3D priors \cite{spatialpin},
program synthesis~\cite{vadar}, or tool-equipped agents~\cite{spatialscore}. GeoJeV selects 3D primitives and their arguments using linear heads on the MLLM's question state, then adds their evidence to its option scores, without generating programs or calling a second language model.

%% file: sec/3_benchmark.tex
\section{\bench}
\label{sec:bench}

\subsection{Overview}
\label{sec:overview}

We introduce \bench to measure the 3D spatial intelligence of MLLMs in hospital rooms from egocentric video. It has
3,368 multiple-choice questions on 569 ten-second clips of 373 layouts in 16 room families. Scenes are generated in
simulation from hospital design standards and every answer is computed from the scene record; the clips are
transferred to photorealistic appearance, verified against the scene and screened by physicians. A disjoint training
set, \trainset, has 44,750 questions.

\begin{figure*}[t]
\centering
\includegraphics[width=\textwidth]{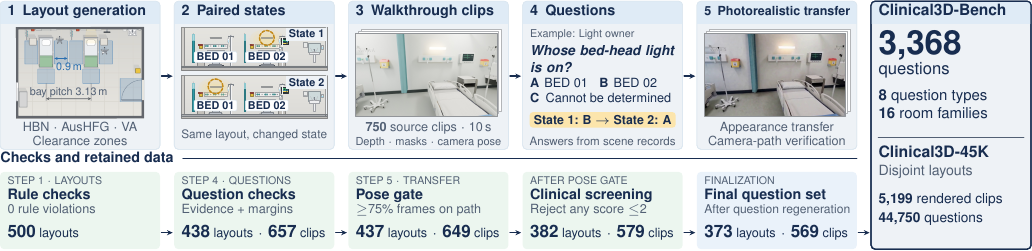}
\caption{\textbf{Construction of \bench.} Top: construction with a paired-state example (abridged). Blue marks
clearance-check regions; bedside/foot zones allow movable equipment. Bottom: checks and retained counts.
Regenerating passage-width questions leaves nine screened layouts without questions, giving the final 373 layouts and 569 clips.}
\label{fig:pipeline}
\end{figure*}

\paragraph{Tasks.}
The eight question types form four task groups (\cref{fig:examples}; shares in \cref{fig:stats}a).
\emph{Spatial memory}: Count asks how many instances of a category seen so far are out of view at a given time, and
Re-ID asks which instance at a later time is the one named earlier. \emph{Ownership and relations}: Light owner asks
which bed the lit bed-head light belongs to, and Device reference asks what a device is mounted on or which label it
carries, where the device is named by its relation to a bed or landmark and a look-alike is in view. \emph{Frames of
reference}: Patient frame asks whether an object is on the left or right of a patient on a named bed; no people are present,
so the model imagines a supine patient, and left and right are the patient's.
\emph{Metric navigation}: Direction asks for the first change of direction on the shortest walkable route from the
camera's start pose to a goal, or for the goal's quadrant; Route asks to fill in the turn sequence of that route or for
the goal nearest by walking; Passability asks whether a passage wide enough for a crash cart (0.9\,m) or transport bed (1.2\,m) reaches the goal. The types comprise 15 subtasks (\cref{tab:subtasks}).

\paragraph{Construction.}
\bench is built in five steps (\cref{fig:pipeline}): scenes are generated from hospital design standards
(\cref{sec:generation}), questions are generated and checked (\cref{sec:qc}), clips are transferred
to photorealistic appearance (\cref{sec:photoreal}), and physicians screen the layouts (\cref{sec:stats}).

\subsection{Standards-grounded scene generation}
\label{sec:generation}

\paragraph{From standards to rules.}
Hospital planning standards state how large a room must be and how much free space its beds need. We take these
requirements for each room type from three public standards (UK Health Building Notes, Australasian Health Facility
Guidelines, US VA Space Planning Criteria), recording the source section of each, and use them in three ways. Size
requirements, such as room area and the spacing between beds, set the ranges from which a layout's dimensions are
sampled. Clearance requirements become geometric checks: a clear strip of 0.9\,m beside each ward bed (1.5\,m in ICU),
a clear zone at the foot of the bed and a free zone at the door. Functional requirements become checks on content and
access: every bed position has a service panel and the required pumps, the room has a handwash station, drawn curtains
do not block the entry, and every bed foot can be reached from the door. Each of the 16 room families has a generator
program~\cite{procthor,infinigenindoors,holodeck}, and a layout is kept only if it passes every check
(\cref{fig:pipeline}); random seeds are fixed, so every layout can be regenerated.

\paragraph{Assets, rendering and clips.}
Free-standing equipment (beds, IV poles, patient monitors, ventilators, imaging and surgical units, chairs, carts;
22 of 47 object categories) uses NVIDIA SimReady assets~\cite{simready} (CC BY 4.0). Everything else is built
procedurally: the parts that questions depend on (infusion pumps with bed labels, bed-head panels, surgical lights
and ceiling pendants, curtains, doors, bed labels), equipment with no suitable SimReady model (CT, stretcher, sharps
containers, sanitiser dispensers) and other furniture (\cref{supp:assets}); small wall items are placed by clinical rules. A camera then walks through the room at eye height (1.55\,m, 90\textdegree{}
horizontal field of view) along a path that passes every bed and comes close enough to each label to read it; it moves
at 0.4--0.9\,m/s and turns at most 30\textdegree/s. The walk is path-traced in Blender at 768$\times$448 and 6\,fps,
and every frame is stored with metric depth, instance and part masks, and the camera intrinsics and pose. From each
walk we keep one 10\,s clip (60 frames), the window in which the most candidate questions can be answered; all
questions are asked on this clip.

\paragraph{Paired states.}
A layout is built in up to three states that differ in one clinical fact: the lit bed-head light
(in corridors, the lit nurse-call light) moves to another bed or room, the privacy curtains change, or corridor items
move. All states of a layout share one clip window, and each state is one clip. In the final set 111 of the 373
layouts keep two or three states. Questions instantiated from the same template on the states of one
layout form a \emph{paired group} of two or three questions; the final set has 262 groups, and in 204 of them the
correct answer changes.

\subsection{Question generation and quality control}
\label{sec:qc}

Questions are instantiated from templates over the scene record. Three groups of checks, all computed from the rendered ground truth, make them answerable and
unambiguous (thresholds in the supplement). \textbf{Evidence:} every object, part or label a question depends on must be
visible, above a pixel threshold and for at least 1\,s, in the 32 frames a model receives. \textbf{Margins:} referring
expressions must pick out one object by a clear distance margin, and near-boundary questions are removed, such as
near-ties in walking distance and passability answers that do not hold 0.1\,m either side of the stated width.
\textbf{Shortcuts:} answer values are balanced within each subtask, and a blind filter~\cite{tst} of two text-only
models and a text-statistics test flags 739 of 3,368 questions (21.9\%); they stay in the official score, and the
\emph{debiased} score excludes them.

\textbf{Model-in-the-loop refinement:} subtask formulations that zero-shot or blind models answered near ceiling were
replaced by harder ones. Refinement acted on formulations, not on individual questions, but since it used Qwen3-VL-8B
and -32B, their scores on \bench may be biased downward relative to other models.

\subsection{Photorealistic clips}
\label{sec:photoreal}

Renders are geometrically exact but look synthetic, while generative video transfer~\cite{cosmostransfer} looks real but carries no 3D ground truth.
We use generators for appearance only and verify their output against the scene. Keyframes are edited with an
image model (gpt-image-2~\cite{gptimage2}), and the render plus edited keyframes condition a video-to-video model
(MiniMax-H3~\cite{minimaxh3}). Each clip is registered to its render by feature
matching~\cite{superpoint,lightglue} and PnP on the rendered depth, and passes if $\geq$75\% of its frames lie on the
source path; 649 of 657 clips pass, with median pose error 0.03\,m and 0.5\textdegree{} (supplement). The gate verifies
the camera path, not that every evidence object survives the transfer: the questions inherit the render-side evidence
checks, and the physician screen removes layouts with visible artefacts. All results use the photorealistic clips; training clips are
renders.

\subsection{Clinical screening and final set}
\label{sec:stats}

\begin{figure}[t]
\centering
\includegraphics[width=\linewidth]{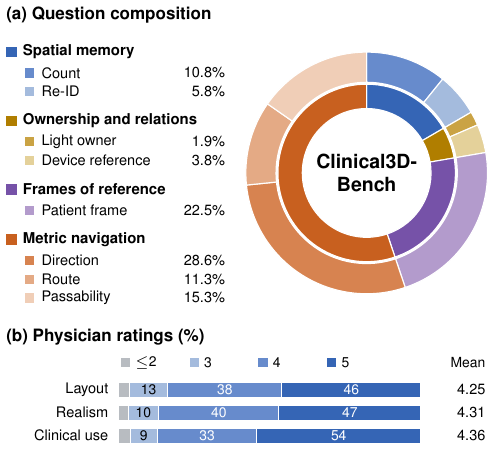}
\caption{\textbf{Statistics of \bench.} (a)~Question shares by task group (inner ring) and type (outer ring).
(b)~Score distributions and means from 874 physician ratings of 437 layouts before screening;
any score $\leq$2 excludes the layout.}
\label{fig:stats}
\end{figure}

\paragraph{Screening.}
Ten resident physicians took part. They first reviewed each question type, and we revised the wording where they
asked, checking that no answer changed. They then rated one photorealistic clip of every layout that passed the pose
gate (437 clips, two physicians each) on three 0--5 scales: layout plausibility, realism and clinical usability (means
4.25 / 4.31 / 4.36; \cref{fig:stats}b). On each scale, the two physicians' scores differ by at most one point for
82--87\% of the layouts. We use the ratings only to remove layouts: a layout is dropped, with all its clips and states,
if either physician scores it 2 or lower on any scale, which removed 55 of the 437 rated layouts.

\paragraph{Final set and training set.}
Of 500 generated layouts (750 clips), 438 have a question that passes the checks, 437 pass the pose gate and 382 pass
screening; the final benchmark keeps 373 (\cref{fig:pipeline}; funnel and composition in the supplement). Each of the
16 families has 16--28 layouts, and every type is asked on changed states (\cref{supp:stats}). With the same chain we
build a training set, \trainset, from 518 further layouts, disjoint from the benchmark's: 44,750 questions on 5,199
clips and a validation set of 2,025 questions. Its clips come with metric depth, camera poses, instance masks and 3D boxes.

%% file: sec/4_evaluation.tex
\section{Evaluating MLLMs}
\label{sec:eval}

\begin{table*}[t]
\centering
\caption{\textbf{Results on \bench} (accuracy \%). Avg.\ is the macro average of the eight types and sets the rank.
\textbf{Bold} and \underline{underline}: best and second best in each column among zero-shot and among fine-tuned models.
$^\dagger$Five physicians on a 400-question subset (50 per type, 80 clips) chosen to match the full set; subset and full-set averages of the 38 models in the table have Spearman $\rho=0.986$.}
\vspace{-0.1cm}
\label{tab:main}
\setlength{\tabcolsep}{4pt}\renewcommand{\arraystretch}{0.9}
\resizebox{\textwidth}{!}{%
\begin{tabular}{l|cccccccc|cc}
\toprule
Methods & Count & Re-ID & Light owner & Patient frame & Device ref. & Direction & Route & Passability & Avg. & Rank \\
\hline
\grp{References}
Chance (analytic) & 20.0 & 33.3 & 28.9 & 50.0 & 27.9 & 24.9 & 26.7 & 50.0 & 32.7 & -- \\
Answer prior (test-fitted) & 34.0 & 39.5 & 46.0 & 54.8 & 39.1 & 28.3 & 30.0 & 54.3 & 40.8 & -- \\
Human$^\dagger$ & 84.0 & 80.0 & 86.0 & 94.0 & 82.0 & 82.0 & 60.0 & 64.0 & 79.0 & -- \\
\hline
\grp{Proprietary Models (API)}
Gemini 3.6 Flash\fr{10} & \textbf{59.2} & 47.2 & 76.2 & 46.0 & \textbf{64.8} & 49.1 & 31.1 & 55.0 & \textbf{53.6} & 4 \\
GPT-5.5\fr{10} & 47.1 & \underline{49.2} & \underline{77.8} & 40.7 & 51.6 & \underline{49.9} & \textbf{37.1} & \underline{57.8} & \underline{51.4} & 5 \\
Claude Sonnet 5.5\fr{10} & 47.1 & \textbf{56.4} & 65.1 & \underline{51.0} & 53.1 & \textbf{51.0} & 31.8 & 55.0 & 51.3 & 6 \\
\hline
\grp{Open-source General Models}
\videojev-4B~\cite{videojev}\fr{20} & 46.6 & 38.5 & 74.6 & 17.0 & 51.6 & 48.4 & 29.5 & \textbf{57.9} & 45.5 & 9 \\
Qwen2.5-VL-3B~\cite{qwen25vl} & 29.0 & 26.7 & 68.3 & 33.3 & 46.1 & 35.2 & 28.4 & 51.9 & 39.9 & 20 \\
Qwen2.5-VL-7B~\cite{qwen25vl} & 17.5 & 40.5 & 17.5 & 27.7 & 41.4 & 26.7 & 30.8 & 54.3 & 32.1 & 35 \\
Qwen2.5-VL-32B~\cite{qwen25vl} & 46.8 & 34.9 & 15.9 & 31.6 & 43.0 & 38.9 & 34.5 & 53.7 & 37.4 & 25 \\
Qwen2.5-VL-72B~\cite{qwen25vl} & 41.4 & 39.0 & 19.0 & 29.0 & 49.2 & 40.4 & 33.4 & 54.3 & 38.2 & 23 \\
Qwen3-VL-4B~\cite{qwen3vl}\fr{16} & 40.5 & 32.8 & 28.6 & 27.1 & 44.5 & 37.2 & 31.3 & 51.0 & 36.6 & 26 \\
Qwen3-VL-8B~\cite{qwen3vl} & 46.6 & 39.0 & 55.6 & 20.9 & 50.0 & 40.0 & 33.4 & 56.0 & 42.7 & 14 \\
Qwen3-VL-32B~\cite{qwen3vl} & 51.2 & 41.0 & 76.2 & 13.4 & 54.7 & 35.9 & 32.9 & 55.2 & 45.1 & 10 \\
InternVL3-8B~\cite{internvl3} & 44.9 & 33.8 & 74.6 & 25.3 & 43.0 & 33.1 & 23.7 & 54.7 & 41.6 & 16 \\
InternVL3-14B~\cite{internvl3} & 52.6 & 47.2 & 71.4 & 20.3 & 47.7 & 40.3 & 31.6 & 54.3 & 45.7 & 7 \\
InternVL3-38B~\cite{internvl3} & 54.2 & 44.1 & 57.1 & 18.6 & \underline{56.2} & 45.2 & 30.3 & 54.7 & 45.1 & 11 \\
InternVL3-78B~\cite{internvl3} & \underline{56.7} & 26.7 & 73.0 & 18.2 & 51.6 & 43.3 & 27.9 & 54.1 & 43.9 & 13 \\
InternVL3.5-4B~\cite{internvl35} & 29.0 & 40.0 & 74.6 & 27.9 & 43.8 & 34.0 & 30.8 & 49.2 & 41.2 & 19 \\
InternVL3.5-8B~\cite{internvl35} & 38.9 & 33.8 & \textbf{79.4} & 36.1 & 44.5 & 37.3 & 28.4 & 54.3 & 44.1 & 12 \\
InternVL3.5-14B~\cite{internvl35} & 47.4 & 32.3 & 71.4 & 32.7 & 50.8 & 41.8 & 34.2 & 54.7 & 45.7 & 8 \\
InternVL3.5-38B~\cite{internvl35} & 47.9 & 31.8 & 60.3 & 21.6 & 49.2 & 44.5 & 28.7 & 54.8 & 42.3 & 15 \\
LLaVA-CoT-11B~\cite{llavacot}\fr{16} & 4.9 & 34.9 & 20.6 & 44.4 & 40.6 & 25.6 & 28.9 & 55.2 & 31.9 & 36 \\
LLaVA-NeXT-Video-34B~\cite{llavanextvideo}\fr{16} & 33.7 & 31.3 & 58.7 & 50.6 & 44.5 & 26.7 & 29.7 & 56.2 & 41.4 & 18 \\
Kimi-VL-A3B-Instruct~\cite{kimivl} & 23.0 & 29.7 & 46.0 & 32.0 & 35.9 & 35.6 & 28.4 & 54.3 & 35.6 & 28 \\
Kimi-VL-A3B-Thinking~\cite{kimivl} & 37.3 & 32.3 & 41.3 & 39.7 & 43.0 & 32.6 & 31.1 & 54.3 & 39.0 & 21 \\
DeepSeek-VL2~\cite{deepseekvl2}\fr{7} & 32.3 & 30.8 & 54.0 & 33.5 & 41.4 & 32.1 & 28.4 & 54.3 & 38.4 & 22 \\
Llama-3.2-11B-Vision~\cite{llama32}\fr{16} & 19.7 & 32.3 & 49.2 & 44.9 & 35.2 & 24.1 & 27.9 & 55.4 & 36.1 & 27 \\
MiMo-VL-7B-SFT~\cite{mimovl} & 34.0 & 32.8 & 9.5 & 24.6 & 36.7 & 41.0 & 29.2 & 54.3 & 32.8 & 33 \\
MiMo-VL-7B-RL~\cite{mimovl} & 26.0 & 30.8 & 23.8 & 25.8 & 35.9 & 42.7 & 31.6 & 53.7 & 33.8 & 31 \\
\hline
\grp{Open-source Spatial Models}
SenseNova-SI-1.1-Qwen2.5-VL-7B~\cite{sensenovasi} & 19.2 & 34.4 & 1.6 & 35.2 & 40.6 & 33.0 & 31.8 & 53.9 & 31.2 & 37 \\
SenseNova-SI-1.1-Qwen3-VL-8B~\cite{sensenovasi} & 41.6 & 43.6 & 4.8 & 25.6 & 38.3 & 32.4 & \underline{35.0} & 56.4 & 34.7 & 30 \\
SenseNova-SI-1.3-InternVL3-8B~\cite{sensenovasi} & 35.6 & 41.0 & 65.1 & 20.3 & 43.0 & 42.2 & 30.8 & 54.3 & 41.5 & 17 \\
Spatial-MLLM-135K~\cite{spatialmllm} & 34.0 & 33.8 & 38.1 & 41.6 & 38.3 & 34.5 & 30.3 & 51.6 & 37.8 & 24 \\
Spatial-MLLM-820K~\cite{spatialmllm} & 20.3 & 32.8 & 52.4 & 32.5 & 39.1 & 24.5 & 29.5 & 53.9 & 35.6 & 29 \\
SpaceR-7B~\cite{spacer} & 17.8 & 36.9 & 20.6 & 29.2 & 43.0 & 31.7 & 33.4 & 55.2 & 33.5 & 32 \\
VST-7B-SFT~\cite{vst} & 24.1 & 29.7 & 6.3 & 27.1 & 46.9 & 36.3 & 33.9 & 53.9 & 32.3 & 34 \\
SpaceLLaVA-13B~\cite{spatialvlm}\fr{8} & 11.2 & 31.3 & 1.6 & \textbf{54.5} & 40.6 & 21.5 & 29.7 & 51.7 & 30.3 & 38 \\
\hline
\grp{Fine-tuned on \trainset}
Spatial-MLLM SFT\fr{16} & 54.8 & 65.6 & \textbf{87.3} & 89.3 & 59.4 & 77.4 & \textbf{65.8} & 59.5 & 69.9 & 3 \\
\videojev-4B SFT\fr{20} & \underline{66.8} & \underline{73.9} & \underline{83.1} & \underline{93.7} & \underline{71.6} & \underline{77.8} & 58.3 & \underline{66.0} & \underline{73.9} & 2 \\
\textbf{\geojev (ours)}\fr{20} & \textbf{69.3} & \textbf{77.8} & 82.5 & \textbf{93.9} & \textbf{71.9} & \textbf{81.2} & \underline{59.5} & \textbf{69.8} & \textbf{75.7} & 1 \\
\bottomrule
\end{tabular}}
\end{table*}

\subsection{Protocol}
\label{sec:protocol}

A model receives the question, options and 32 frames sampled uniformly from the 60 of a clip (or the number
beside its name in \cref{tab:main}), each preceded by its timestamp, and answers with one option. Accuracy per type pools its subtasks by question; the
headline is the \textbf{macro average} over the eight types (chance 32.7). As references we report uniform
random choice, text-only runs and an \emph{answer prior}, a constant-answer baseline that picks the most frequent answer
of each subtask, fitted on the test labels (40.8 macro).

\subsection{Results}
\label{sec:main}

\begin{figure*}[t]
\centering
\includegraphics[width=\textwidth]{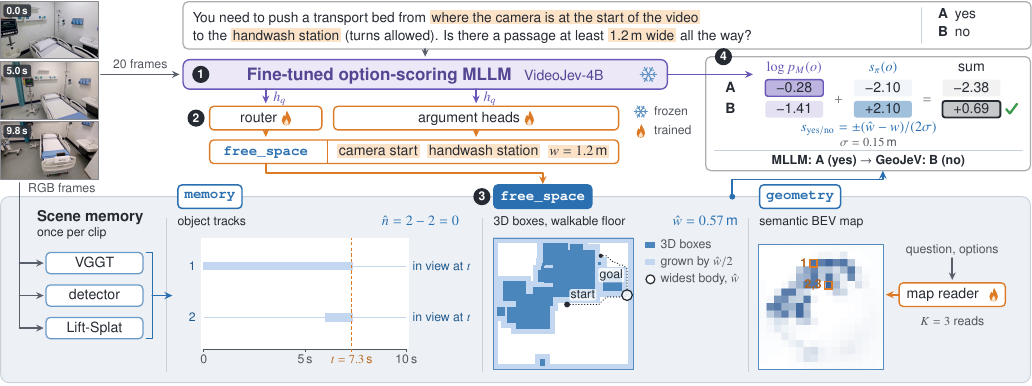}
\caption{\textbf{\geojev} on a passability question. \videojev-4B, fine-tuned and then frozen, calls \texttt{free\_space}, which estimates a maximum traversable body width of 0.57\,m, below the required 1.2\,m. Adding its scores to the MLLM's log-probabilities corrects ``yes'' to ``no''. \texttt{memory} and \texttt{geometry} show separate calls on the same clip.}
\label{fig:method}
\end{figure*}

\paragraph{Zero-shot models are close to the answer prior.}
The three proprietary models score 51.3--53.6 (\cref{tab:main}), 6--8 points above the best open model, but 20 or
more below the fine-tuned \videojev-4B; none is clearly above chance on Patient frame (40.7--51.0), and Route stays at
31.1--37.1. The 32 open MLLMs score 30.3--45.7. Thirteen are above the
test-fitted answer prior (40.8), and the best two (InternVL3-14B and InternVL3.5-14B, both 45.7) are 5 points above it. Removing
blind-flagged questions changes little (Qwen3-VL-32B 43.3, \videojev-4B 44.1, answer prior 37.3). On Passability every zero-shot model is within 8 points of chance.
The spatial MLLMs, trained on spatial question answering, stay below the best general models, and
within a family a larger model is not reliably better. Text-only models stay near or below the answer prior on every type.

\paragraph{Physicians.}
Five physicians score 79.0 on a 400-question subset, 25.4 points above the best zero-shot model. They reach 94.0 on
Patient frame, where no zero-shot model is clearly above chance, and 80--86 on five other types, but 60.0 on Route
and 64.0 on Passability: the two types weakest for fine-tuned \videojev-4B are also hardest for people.

\paragraph{Perspective taking is the clearest failure.}
On the two-way Patient frame, 27 of the 32 open zero-shot models score below 40\% and none is clearly above the 50\% chance (at most 54.5). Beds are mostly seen from the foot, where the patient's left is the viewer's right: in all 241 bedside items where the
object has a side of the bed in the image, that side is opposite to the patient's. Strong open models are among the worst:
Qwen3-VL-32B scores 13.4 and gives the image-side answer on 93.4\% of these 241 items (its \emph{mirrored-answer
rate}). The failure is not confined to these: zero-shot models score 17--53\% on the room-scale items.

\paragraph{Paired states expose reliance on layout priors.}
In the 204 paired groups whose correct answer changes between states, a model that picks the same option in every
state is wrong at least once. Qwen3-VL-32B does so in 42.2\% of the groups, 15.4 points more than its answer
preferences alone give (the rate when each question is replaced by one of the same subtask from another layout). This
excess is 3.1--41.1 points (median 14.8) for open zero-shot models and 4.1--9.3 for proprietary ones, against $-$0.5
to +4.3 for the fine-tuned \videojev-4B (supplement).

\paragraph{Fine-tuning helps; planning lags.}
Fine-tuning on \trainset lifts \videojev-4B from 45.5 to
73.9 (three seeds), yet Route (58.3) and Passability (66.0) remain its weakest types, while Patient frame
rises from 17.0 to 93.7. Spatial-MLLM, pre-trained on 3D spatial question answering, has the best Route after
fine-tuning (65.8), but overall it stays 4.0 points below \videojev-4B SFT. This is consistent with 3D pre-training helping routes after
fine-tuning. Training and test layouts are disjoint but share a generator, and training
clips are renders, so the gains hold across the appearance change.

%% file: sec/5_geojev.tex
\section{Making MLLMs use 3D: \geojev}
\label{sec:geojev}

\subsection{Is 3D given as input used?}
\label{sec:shuffle}

\paragraph{Shuffle twins.}
For a 3D-augmented model we evaluate a twin whose 3D comes from another clip at test time; the model uses the scene's
3D only if its twin is clearly worse. We test four ways of giving a fine-tuned MLLM 3D (\cref{tab:shuffle}): Lift-Splat~\cite{liftsplat} BEV
tokens appended to the visual tokens, local BEV reads injected into the language model, a geometry head on its hidden
states, and, as a privileged input, ground-truth BEV maps. With BEV tokens, local reads or the ground-truth map, the
twin is within 0.5 points of the real model. The geometry head does read the scene (1.4 points lower with another
clip's 3D) but adds only 0.6 points over the plain fine-tune. No 3D input thus lifts the model clearly above it, and
\geojev instead adds 3D as computation outside the MLLM.

\begin{table}[t]
\centering
\caption{\textbf{Shuffle twins.} Macro accuracy (\%, three seeds) of \videojev-4B fine-tuned with 3D input, given its
own clip's 3D (Own) or another clip's (Shuf.), on the benchmark and on its source renders. $\Delta$ = Own $-$ Shuf.,
with 95\% CI.}
\label{tab:shuffle}
\setlength{\tabcolsep}{2.4pt}
\resizebox{\columnwidth}{!}{%
\begin{tabular}{@{}lcccccc@{}}
\toprule
 & \multicolumn{3}{c}{Benchmark} & \multicolumn{3}{c}{Renders} \\
\cmidrule(lr){2-4}\cmidrule(l){5-7}
3D input & Own & Shuf. & $\Delta$ & Own & Shuf. & $\Delta$ \\
\midrule
None (\videojev-4B SFT) & 73.9 & -- & -- & 77.2 & -- & -- \\
Lift-Splat BEV tokens & 73.2 & 73.0 & +0.2 {\scriptsize[$-$0.4, +0.7]} & 75.8 & 75.8 & 0.0 {\scriptsize[$-$0.5, +0.5]} \\
Local BEV reads & 74.2 & 74.1 & +0.1 {\scriptsize[$-$0.2, +0.4]} & 76.4 & 76.2 & +0.2 {\scriptsize[$-$0.2, +0.6]} \\
Geometry head on hidden states & 74.5 & 73.1 & +1.4 {\scriptsize[+0.9, +1.9]} & 76.4 & 75.0 & +1.4 {\scriptsize[+0.9, +2.0]} \\
GT BEV at GT positions & 72.9 & 73.3 & $-$0.4 {\scriptsize[$-$0.9, +0.1]} & 76.0 & 75.5 & +0.5 {\scriptsize[+0.1, +1.0]} \\
\bottomrule
\end{tabular}}
\end{table}

\subsection{\geojev}
\label{sec:geojev_method}

\geojev (\cref{fig:method}) has three parts: a fine-tuned MLLM that scores each option, a \emph{scene memory}
reconstructed once per clip, and three \emph{primitives} that compute on it and are shared across question types. The
MLLM decides from its own hidden state which primitive to call and with which arguments; it runs once per question and
is not changed.

\paragraph{Option-scoring MLLM.}
\geojev builds on \videojev-4B~\cite{videojev}, which encodes the question and each option and scores
option $o$ with a cosine decision head on its final hidden states,
$\ell(o) = s \cdot \cos(W_q h_q, W_o h_o) + b$, and $\pM(o \mid V, q) = \softmax_o \ell(o)$ for
frames $V$ and question $q$. Because each option has its own score, 3D evidence about an option can be added to that
option directly. The model is fine-tuned on \trainset and then frozen.

\paragraph{Scene memory.}
Each clip is reconstructed once from its RGB frames: VGGT~\cite{vggt} gives depth and camera poses, a
detector~\cite{yolo} gives objects, which are lifted to 3D and linked into tracks, and a Lift-Splat
encoder~\cite{liftsplat} gives a semantic bird's-eye-view (BEV) map and 3D object boxes. The memory holds the tracks
with their time stamps, the boxes, the walkable floor and the BEV map; no benchmark depth, pose, mask or label is used.

\paragraph{Three primitives.}
\texttt{memory} looks up the tracks of one class: how many are out of view at time $t$, how many there are in all,
or which detection at $t'$ is the same object as the left- or rightmost one at $t$. \texttt{free\_space} searches the
walkable floor around the object boxes for the widest body that still gets from a start to a goal.
\texttt{geometry}, which serves the remaining spatial questions, has no arguments: a learned \emph{map reader} takes
the question and option text, makes a few hard reads of cells of the BEV map and scores each option from them. It is
trained by on-policy distillation~\cite{dagger,gkd,minillm} from a privileged teacher~\cite{lbc} that reads
ground-truth maps. \texttt{memory} and \texttt{free\_space} take arguments (an object class, times, a start and a
goal, a width), never the question's wording. What they compute follows from the answer format of the options, not
from the question type: numbers ask for a count, ordinal positions among objects of one class for an identity and
``yes''/``no'' for a width test. Each primitive returns a score per option, $s(o)$: $-|k-\hat n|$ for count option $k$, a
log-softmax over 3D distances for identity, $\pm(\hat w - w)/2\sigma$ for ``yes''/``no'' at width $w$, and the map
reader's log-softmax for geometry.

\paragraph{The MLLM calls the primitives.}
Let $h_q$ be the MLLM's question state, the decision head's query input. Linear heads on $h_q$ give the primitive,
$\pi = \mathrm{route}(h_q)$, and its arguments $\theta = \mathrm{args}(h_q)$: object classes, the side of the
reference, and which number of the question is a time or a width. No head predicts the question type or the operation.
\geojev answers
\begin{equation}
\hat o = \argmax_o \big[\log \pM(o \mid V, q) + s_{\pi}(o;\, \theta)\big],
\label{eq:combine}
\end{equation}
and when the chosen primitive cannot run (an argument not found in the scene memory, or no option format it computes)
$s_\pi = 0$ and the MLLM's own answer is kept. The router needs no question-type labels: its target is the set of
primitives that alone answer a training question correctly, so a question goes where 3D computation helps. No second
language model is involved, and the heads add no forward pass (details of all parts in the supplement).

\subsection{Results of \geojev}
\label{sec:geojev_results}

\paragraph{\geojev improves the fine-tuned MLLM.}
Over three fine-tuning seeds, \geojev raises the macro from 73.9 to 75.7, each seed with its own heads (+1.8, paired layout-bootstrap 95\% CI [+0.7, +3.0]; every
seed's gain is above zero). The gains sit on the types the primitives target (\cref{tab:main}): Count and Re-ID
(\texttt{memory}, +2.5 and +3.9), Passability (\texttt{free\_space}, +3.8) and Direction (\texttt{geometry}, +3.4);
the other types change by 1.2 points or less.

\begin{table}[t]
\centering
\caption{\textbf{Ablations} (3-seed macro \%). All rows use the same fine-tuned
\videojev-4B; a removed primitive's questions keep the MLLM's own answer.}
\label{tab:ablation}
\footnotesize
\begin{tabular}{@{}lc@{}}
\toprule
 & Macro \\
\midrule
\videojev-4B SFT & 73.9 \\
\textbf{\geojev} & \textbf{75.7} \\
\midrule
$-$ \texttt{memory} & 75.0 \\
$-$ \texttt{free\_space} & 75.4 \\
$-$ \texttt{geometry} & 75.0 \\
primitives given another layout's scene memory & 70.4 \\
\texttt{geometry} given another layout's BEV map & 74.4 \\
\bottomrule
\end{tabular}
\end{table}

\paragraph{Each primitive contributes, and the scene's 3D is used.}
Removing any one primitive costs 0.3--0.7 points, each on its own types (\cref{tab:ablation}). Given another layout's
scene memory (same room family, same state), \geojev falls to 70.4, 3.5 points below the MLLM alone, with Re-ID at 55.7 instead of 77.8 and Count at 62.1 instead of 69.3: unlike the 3D
inputs of \cref{sec:shuffle}, the evidence it adds is specific to the scene. This holds for the BEV map alone: given
another layout's map, with the other primitives unchanged, the map reader lowers the macro by 1.3 points (paired 95\% CI
[+0.6, +2.1]) to 74.4, below removing \texttt{geometry} altogether, and Direction falls from 81.2 to 76.1.

%% file: sec/6_conclusion.tex
\section{Conclusion}
\label{sec:conclusion}

\bench measures 3D spatial intelligence in hospital rooms with exact, scene-derived answers,
answer-changing scene states and pose-verified photorealistic clips. Zero-shot MLLMs stay close to the answer prior
and fail at the patient's frame of reference; fine-tuning supplies that convention but leaves route planning and
passability the weakest types. 3D added to the fine-tuned model as input is left largely unused, whereas \geojev lets an option-scoring
MLLM call 3D primitives from its own hidden state and add their evidence to its per-option scores.

%% file: sec/X_suppl.tex
\section{Scene assets}
\label{supp:assets}

Of the 47 object categories in the benchmark scenes, 22 use NVIDIA SimReady assets~\cite{simready} and 25 are built
procedurally (\cref{tab:assets}). We place each SimReady model in the same local frame as the procedural meshes it replaces. We keep the procedural parts that questions refer to (\eg bed and chair number plates) on top of the model.

\begin{table*}[t]
\centering
\caption{\textbf{Scene assets.} Source of each of the 47 object categories (structural walls and floors excluded).}
\label{tab:assets}
\small
\begin{tabular}{@{}p{0.27\linewidth}p{0.69\linewidth}@{}}
\toprule
Source & Categories \\
\midrule
SimReady (22) & hospital bed; bedside cabinet; chair; IV pole; patient monitor (floor stand); glove box; handwash
station; infusion chair; phlebotomy chair; exam stool; suction unit; ventilator; resuscitation cart; computer cart;
wheelchair; workstation display; X-ray table, tube stand and detector stand; operating table; anaesthesia workstation;
ultrasound console \\
\midrule
Procedural: parts questions refer to (9) & infusion pump (bed label, pole clamp); bed-head service panel (lights);
oxygen flowmeter; privacy curtain (open/closed); room door (open/closed); monitor wall mount; surgical light head;
ceiling light support; ceiling service pendant \\
\midrule
Procedural: no suitable SimReady model (6) & CT gantry; CT table; resuscitation stretcher; sharps container;
hand-sanitiser dispenser; privacy screen \\
\midrule
Procedural: other furniture (10) & storage cabinet; equipment shelving; service counter; instrument cart; specimen
trolley; examination bed; endoscopy tower; entry partition; waste bin; supply container \\
\bottomrule
\end{tabular}
\end{table*}

\section{Question definitions and quality-control thresholds}
\label{supp:qc}

\paragraph{Types and subtasks.}
A subtask is a single question form for which we balance answer values separately. We group the 15 subtasks into the eight question types used in the main paper (\cref{tab:subtasks}). Accuracy per type pools the questions of its subtasks.

\begin{table*}[t]
\centering
\caption{\textbf{The 15 subtasks of \bench} and the eight question types they are reported as. \#Q is the number of
questions in the final set; Opt.\ is the number of answer options. Other tables abbreviate Light owner,
Patient frame, Device reference and Passability as Light, Patient, Device and Pass.}
\label{tab:subtasks}
\footnotesize
\setlength{\tabcolsep}{5pt}
\begin{tabular}{@{}llp{0.56\textwidth}rc@{}}
\toprule
Type & Subtask & The question asks & \#Q & Opt. \\
\midrule
Count & Out-of-view count & how many instances of a category seen so far are out of view at time $t$ & 365 & 5 \\
\midrule
Re-ID & Re-identification & which instance at $t'$ is the leftmost one of its category at $t$ & 195 & 3--4 \\
\midrule
Light owner & Lit-light owner & which bed the lit bed-head light belongs to & 63 & 3--7 \\
\midrule
Patient frame & Bedside & whether an object within 1.5\,m of the bed is on the supine patient's left or right & 461 & 2 \\
 & Room scale & the same for an object elsewhere in the room & 298 & 2 \\
\midrule
Device ref. & Mount & what the device nearest a named landmark is mounted on & 110 & 3--4 \\
 & Label & which bed label is written on the device nearest a named landmark & 8 & 3--4 \\
 & Label side & beside which bed, and on which side of its patient, the device with a given label is & 10 & 4 \\
\midrule
Direction & First turn & the first change of direction on the shortest walkable route to the goal & 349 & 4 \\
 & Quadrant & the goal's quadrant from the start pose & 587 & 4 \\
 & Door position & which corridor door, counted from the entrance on each side, is the goal & 26 & 4--6 \\
\midrule
Route & Route plan & the missing turns in the turn sequence of the shortest route & 314 & 4 \\
 & Nearest by walking & which of several goals is nearest along the walking route & 66 & 2--4 \\
\midrule
Passability & Passage width & whether a passage at least 0.9\,m (crash cart) or 1.2\,m (transport bed) wide leads to the goal & 459 & 2 \\
 & Straight route & whether a transport bed of the stated size (1.0\,m wide), kept at least 0.02\,m clear on both sides and above, can be pushed along a straight route marked on the floor (a fixed path, unlike the passage widths above) & 57 & 2 \\
\bottomrule
\end{tabular}
\end{table*}

\paragraph{Navigation.}
Routes are shortest paths on a 0.1\,m occupancy grid with 0.35\,m body clearance. Walking distance is measured to the
nearest standable cell beside the goal. Direction changes under 30\textdegree{} are ignored and changes over
135\textdegree{} count as turning around.
The starting point is the camera's pose in the first frame or the foot of a labelled bed. Route plans can also start next to a named object, with the walker facing another. Goals are clinical cues: the bed whose bed-head light is on, a unique device
such as a crash cart, handwash station or patient monitor, a bed by its plate, or the room whose nurse-call light is on.

\paragraph{Distractors.}
Light owner offers ``cannot be determined'' as a distractor. In Mount and Label, a look-alike device of the same class is in view, so the model must identify the target device by its relation to the named bed or landmark.

\paragraph{Evidence.}
We keep a question only if each evidence item appears in one of the 32 frames a model receives and is visible for at least 1\,s without interruption in the clip (Device reference: the first condition only). The minimum visible areas are 400 pixels for an object, 60 for a part such as a lamp or label, and 90 for text that must be read. Re-ID uses its two named frames instead: the target is the leftmost instance at $t$, and $t'$ shows two or three instances, each of $\geq$1,500 pixels, clear of the image border and 40 pixels apart. If the target is not among them (77 of 195 questions), the answer is ``not in that frame''.
A Count question is dropped if any instance is only partly visible (at least 30 pixels but never reaching the counting
threshold).

\paragraph{Margins.}
Referring expressions must pick out one object. In Mount, the runner-up device must be at least 0.3\,m farther from the landmark than the nearest device and at least 1/0.85 times as far. In Label, it must be at least 0.5\,m farther. The following near-boundary questions are removed:
\begin{itemize}[leftmargin=*,itemsep=1pt,topsep=2pt]
\item quadrant questions within 20\textdegree{} of a quadrant boundary;
\item nearest-by-walking questions won by under 0.6\,m;
\item routes whose turn sequence changes when the body clearance grows by 0.1\,m, or that have legs under 0.8\,m or turns of
  30--45\textdegree;
\item passage-width questions whose widest passable width is not 0.1--0.35\,m from the stated width, and straight routes
  cleared or blocked by under 0.15\,m.
\end{itemize}

\paragraph{Shortcuts.}
When selecting questions, we cap the share of each answer value at $\min(0.55, 1/k + 0.1)$ within its subtask or within a finer group where a cue could reveal the answer (\eg within each wording template). Here $k$ counts the values at least a tenth as frequent as the most common one. In Count, ``4 or more'' falls below that threshold, so $k=4$ and the cap is 35\%. We cap each answer value in Quadrant at 27\%. For passage-width questions, we cap ``yes'' at 55\% for each stated width. Route distractors follow the training-split frequency of turn sequences, and each clip
contributes a capped number of questions per subtask.

\paragraph{Blind filter.}
The blind filter gives the question and options to two text-only models, Qwen3-VL-8B and -32B~\cite{qwen3vl}. It also uses a naive-Bayes text-statistics test cross-validated by layout. The filter flags a question if both models answer it correctly or the test assigns the gold answer a probability of $\geq$0.8. For the passage-width questions only the text-statistics test applies.

\paragraph{Clip selection.}
For each camera path, we select the 10\,s window (60 frames) with the most answerable candidates, summed over the states walked along the path and weighted by $1/\sqrt{\text{subtask candidate count}}$; a candidate counts if its evidence is visible for 1\,s inside the window. States that share a path (usually the first two) share the window; a state with its own path gets its own.

\section{Photorealistic clips: registration and gate}
\label{supp:photoreal}

The generated video does not keep the render's timing, so each clip is registered back
to its render.
We use SuperPoint + LightGlue to match photorealistic frames to source frames within $\pm$3\,s. We back-project the matches using the rendered depth and estimate the camera pose with PnP-RANSAC, using one zoom factor per clip. A frame is \emph{on path} if its pose differs from the source pose by no more than 0.3\,m laterally, 0.6\,m axially and 10\textdegree{} in orientation. An unlocalised gap over 1\,s counts as off path. A clip passes if at least 75\% of its frames are on path, and an order-preserving alignment then assigns one
photorealistic frame to each source frame. Each clip is generated with three keyframes, and a clip that fails the gate is generated once more with five. 649 of
657 clips pass, with median pose error 0.03\,m and 0.5\textdegree. Of the aligned frames, 90.4\% lie within 0.2\,m lateral, 0.4\,m
axial and 5\textdegree{} of the source.

\paragraph{Prompts.}
We use the image model gpt-image-2 to edit the keyframes: three at 0, 5 and 9.7\,s, or five evenly spaced. The prompt asks the model to depict the render of \emph{[room]} (\eg ``an unoccupied multi-bed hospital ward'') as an unretouched photograph of a hospital in daily use. It asks the model to preserve the exact viewpoint and framing, all object counts, positions and outlines, and all visible text. Nothing may be added, removed or moved. We then give MiniMax-H3 (768P, 10\,s) the render as the reference video and the edited keyframes as references for appearance only. The prompt asks for the same shot as real handheld footage. It asks the model to follow the render's camera ``frame for frame'', keep every object and all text, and show no people.

\section{Final set: funnel and composition}
\label{supp:stats}

\paragraph{Funnel.}
500 benchmark layouts (750 clips) $\to$ 438 layouts / 657 clips with at least one question that passes the checks
$\to$ 437 / 649 passing the pose gate $\to$ 382 / 579 after physician screening $\to$ 373 layouts / 569 clips / 3,368
questions in the final set (10 clips from 9 layouts have no questions left after we apply the passage-width margins). Of the
373 layouts, 262 have one state in the final set, 26 have two and 85 have three.

\begin{figure}[t]
\centering
\includegraphics[width=0.9\linewidth]{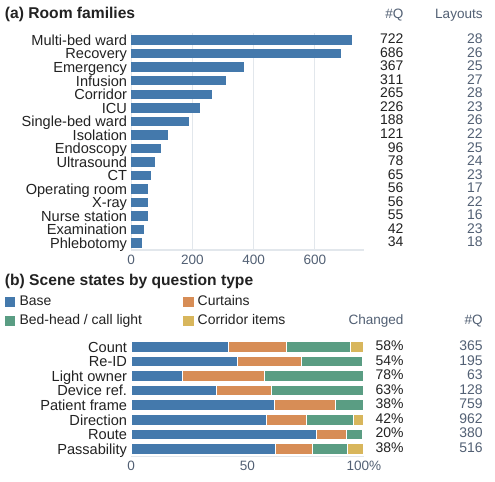}
\caption{\textbf{Room and state composition of \bench.} (a)~Questions (\#Q) and layouts per room family.
(b)~State shares per question type; the right columns give the combined changed-state share and question count.}
\label{fig:composition}
\end{figure}

\paragraph{Composition.}
\Cref{fig:composition} gives question and layout counts per room family and state shares per question type.

\paragraph{Paired groups.}
Questions instantiated from the same template on two or three states of a layout form a paired group. The final set has 262 such groups. In 204 of them the correct answer changes between states.

\paragraph{Training split.}
The 518 further layouts are disjoint from the
benchmark. We render the full walkthroughs without photorealistic transfer or physician screening and tile them into 10\,s windows. We generate and check questions as we do for the benchmark, but without per-clip caps. This gives 44,750 training questions on 5,199 clips from 484 layouts and a validation set of 2,025 questions on 244 clips from 22 layouts. The other 12 layouts yield no question. Four metric subtasks (absolute
distance, object size, relative direction, camera distance) appear only in training.

\paragraph{Fine-tuning recipe.}
For both \videojev-4B SFT and Spatial-MLLM SFT, we freeze the visual encoder and train all language-model weights without adapters for one epoch on the training split. We use AdamW with a learning rate of $10^{-5}$, 3\% warm-up, cosine decay and gradient clipping at 1.0. Each model takes its official video input. \videojev-4B takes frames at 2\,fps (20 per 10\,s clip) at the native resolution of 768$\times$448. Each update uses 8 videos and all their questions. We train its answer-scoring decision head at a learning rate of $10^{-4}$. Spatial-MLLM SFT starts from Spatial-MLLM-v1.1-Instruct-820K and uses its official 16-frame input, a global batch of
48 questions and one seed; \videojev-4B SFT uses three seeds.

\section{Additional evaluation results}
\label{supp:eval}

\paragraph{Protocol details.}
From each clip's 60 frames, the protocol selects the 32 frames with indices $\mathrm{round}(\mathrm{linspace}(0, 59, 32))$. Open models that have a reasoning mode use it for decoding. If a model's context cannot hold 32 frames or the model was fine-tuned with a different frame budget, we use the frame count beside its name in the main table. For \videojev-4B, we use its official input of 20 frames at 2\,fps in both zero-shot and fine-tuned runs. For every model and all 195 Re-ID questions, we check whether the sample includes the two frames named in the question and add any that are missing. The number beside a model's name is therefore its base budget. Bootstrap confidence intervals resample whole layouts. The answer prior is optimistic because we fit it on the test labels. It exceeds the analytic chance of 32.7 because the caps under Shortcuts still allow skewed answer shares.

\paragraph{Physicians.}
Each of the five physicians answered 80 of the 400 questions in the physician subset. They saw the benchmark's question and options, watched the full clip and could replay it freely with no time limit. The five are among the ten who screened the layouts. All took part voluntarily and without pay. We told them how we would use their answers and kept only anonymised answers.

\paragraph{\bench-mini.}
The 400 questions the physicians answered (50 per type, 80 clips, all 16 room families) form \bench-mini, a smaller
set with a human reference. The clips cover the room-family mix of each type with as few clips as possible. Within
them, we drew the questions of each type to match the full set in subtask, answer-letter and blind-flag shares and in
the per-type accuracy of the 25 main-table models evaluated at the time (all but Qwen2.5-VL-3B/32B/72B, Qwen3-VL-4B,
InternVL3-8B/14B/38B, InternVL3.5-4B/8B/14B/38B and \videojev-4B; \geojev in an earlier version) and of runs with other
input settings. \Cref{tab:mini} gives every model of the main table on
\bench-mini under the main-table protocol. Mini and full-set averages differ by 1.3 points on average and by at most
4.3, with Spearman $\rho = 0.986$ over the 38 models; the accuracy matching favours this agreement. Here the physicians are
22.0 points above the best zero-shot model (57.0).

\begin{table*}[t]
\centering
\caption{\textbf{Results on \bench-mini} (400 questions, accuracy \%), laid out as the main table. Avg.\ is the macro
average of the eight types, computed from per-question results and rounded once; it sets the rank. \textbf{Bold} and
\underline{underline}: best and second best in each column among zero-shot and among fine-tuned models. Chance and
answer prior are computed on the mini set itself; \videojev-4B SFT and \geojev are the mean of three seeds, Spatial-MLLM
SFT is one seed.}
\label{tab:mini}
\setlength{\tabcolsep}{4pt}\renewcommand{\arraystretch}{0.9}
\resizebox{\textwidth}{!}{%
\begin{tabular}{l|cccccccc|cc}
\toprule
Methods & Count & Re-ID & Light owner & Patient frame & Device ref. & Direction & Route & Passability & Avg. & Rank \\
\hline
\rowcolor{rowblue}\textit{\textbf{References}} & & & & & & & & & & \\
Chance (analytic) & 20.0 & 33.3 & 29.5 & 50.0 & 28.2 & 24.8 & 27.0 & 50.0 & 32.8 & -- \\
Answer prior (test-fitted) & 34.0 & 40.0 & 50.0 & 54.0 & 40.0 & 28.0 & 34.0 & 62.0 & 42.8 & -- \\
Human & 84.0 & 80.0 & 86.0 & 94.0 & 82.0 & 82.0 & 60.0 & 64.0 & 79.0 & -- \\
\hline
\rowcolor{rowblue}\textit{\textbf{Proprietary Models (API)}} & & & & & & & & & & \\
Gemini 3.6 Flash\fr{10} & \textbf{58.0} & 56.0 & \underline{82.0} & \underline{48.0} & \textbf{78.0} & 50.0 & 28.0 & 56.0 & \textbf{57.0} & 4 \\
GPT-5.5\fr{10} & 46.0 & 50.0 & \textbf{84.0} & 44.0 & 50.0 & \underline{52.0} & \underline{38.0} & \underline{60.0} & \underline{53.0} & 5 \\
Claude Sonnet 5.5\fr{10} & 48.0 & \textbf{60.0} & 66.0 & \textbf{52.0} & 50.0 & \textbf{56.0} & 32.0 & \underline{60.0} & \underline{53.0} & 5 \\
\hline
\rowcolor{rowblue}\textit{\textbf{Open-source General Models}} & & & & & & & & & & \\
\videojev-4B\fr{20} & 52.0 & 54.0 & 74.0 & 16.0 & 56.0 & 50.0 & 24.0 & 58.0 & 48.0 & 8 \\
Qwen2.5-VL-3B & 30.0 & 32.0 & 70.0 & 28.0 & 52.0 & 40.0 & 22.0 & 50.0 & 40.5 & 21 \\
Qwen2.5-VL-7B & 20.0 & 48.0 & 16.0 & 32.0 & 46.0 & 28.0 & 32.0 & 52.0 & 34.3 & 33 \\
Qwen2.5-VL-32B & 48.0 & 34.0 & 16.0 & 36.0 & 46.0 & 36.0 & \textbf{40.0} & 54.0 & 38.8 & 22 \\
Qwen2.5-VL-72B & 40.0 & 34.0 & 22.0 & 28.0 & 54.0 & 32.0 & 34.0 & 54.0 & 37.3 & 24 \\
Qwen3-VL-4B\fr{16} & 36.0 & 44.0 & 28.0 & 28.0 & 46.0 & 40.0 & 34.0 & 38.0 & 36.8 & 28 \\
Qwen3-VL-8B & 44.0 & 40.0 & 56.0 & 18.0 & 52.0 & 40.0 & 28.0 & 56.0 & 41.8 & 19 \\
Qwen3-VL-32B & 50.0 & 38.0 & 80.0 & 12.0 & 62.0 & 28.0 & 32.0 & \underline{60.0} & 45.3 & 12 \\
InternVL3-8B & 38.0 & 36.0 & 74.0 & 24.0 & 44.0 & 42.0 & 28.0 & 54.0 & 42.5 & 17 \\
InternVL3-14B & 50.0 & \underline{58.0} & 74.0 & 26.0 & 56.0 & 46.0 & 36.0 & 54.0 & 50.0 & 7 \\
InternVL3-38B & \textbf{58.0} & 44.0 & 56.0 & 18.0 & \underline{64.0} & 48.0 & \underline{38.0} & 50.0 & 47.0 & 10 \\
InternVL3-78B & \underline{56.0} & 30.0 & 70.0 & 18.0 & 54.0 & 42.0 & 28.0 & 54.0 & 44.0 & 13 \\
InternVL3.5-4B & 24.0 & 52.0 & 80.0 & 30.0 & 46.0 & 32.0 & 36.0 & 42.0 & 42.8 & 16 \\
InternVL3.5-8B & 34.0 & 50.0 & \underline{82.0} & 40.0 & 48.0 & 46.0 & 34.0 & 50.0 & 48.0 & 8 \\
InternVL3.5-14B & 44.0 & 46.0 & 74.0 & 28.0 & 50.0 & 40.0 & 32.0 & 58.0 & 46.5 & 11 \\
InternVL3.5-38B & 44.0 & 36.0 & 68.0 & 22.0 & 54.0 & 44.0 & 24.0 & 56.0 & 43.5 & 14 \\
LLaVA-CoT-11B\fr{16} & 8.0 & 28.0 & 20.0 & 44.0 & 42.0 & 26.0 & 26.0 & 54.0 & 31.0 & 36 \\
LLaVA-NeXT-Video-34B\fr{16} & 32.0 & 42.0 & 60.0 & \underline{48.0} & 40.0 & 26.0 & 32.0 & 56.0 & 42.0 & 18 \\
Kimi-VL-A3B-Instruct & 22.0 & 24.0 & 44.0 & 44.0 & 40.0 & 40.0 & 28.0 & 54.0 & 37.0 & 25 \\
Kimi-VL-A3B-Thinking & 38.0 & 32.0 & 46.0 & 36.0 & 42.0 & 26.0 & 26.0 & 50.0 & 37.0 & 25 \\
DeepSeek-VL2\fr{7} & 32.0 & 40.0 & 60.0 & 34.0 & 46.0 & 34.0 & 28.0 & 54.0 & 41.0 & 20 \\
Llama-3.2-11B-Vision\fr{16} & 18.0 & 36.0 & 46.0 & 46.0 & 42.0 & 20.0 & 32.0 & 56.0 & 37.0 & 25 \\
MiMo-VL-7B-SFT & 32.0 & 32.0 & 10.0 & 28.0 & 40.0 & 38.0 & 34.0 & 52.0 & 33.3 & 35 \\
MiMo-VL-7B-RL & 30.0 & 34.0 & 24.0 & 28.0 & 38.0 & 40.0 & 34.0 & 48.0 & 34.5 & 31 \\
\hline
\rowcolor{rowblue}\textit{\textbf{Open-source Spatial Models}} & & & & & & & & & & \\
SenseNova-SI-1.1-Qwen2.5-VL-7B & 20.0 & 30.0 & 2.0 & 38.0 & 44.0 & 26.0 & 32.0 & 54.0 & 30.8 & 37 \\
SenseNova-SI-1.1-Qwen3-VL-8B & 38.0 & 50.0 & 6.0 & 28.0 & 44.0 & 30.0 & 34.0 & \underline{60.0} & 36.3 & 30 \\
SenseNova-SI-1.3-InternVL3-8B & 36.0 & 50.0 & 70.0 & 16.0 & 44.0 & 44.0 & 32.0 & 54.0 & 43.3 & 15 \\
Spatial-MLLM-135K & 32.0 & 28.0 & 38.0 & 40.0 & 38.0 & 34.0 & 30.0 & \underline{60.0} & 37.5 & 23 \\
Spatial-MLLM-820K & 22.0 & 32.0 & 54.0 & 32.0 & 36.0 & 22.0 & 32.0 & \textbf{64.0} & 36.8 & 28 \\
SpaceR-7B & 20.0 & 34.0 & 20.0 & 32.0 & 50.0 & 32.0 & \underline{38.0} & 50.0 & 34.5 & 31 \\
VST-7B-SFT & 24.0 & 34.0 & 6.0 & 32.0 & 48.0 & 40.0 & 32.0 & 54.0 & 33.8 & 34 \\
SpaceLLaVA-13B\fr{8} & 12.0 & 32.0 & 0.0 & \textbf{52.0} & 44.0 & 24.0 & \underline{38.0} & 44.0 & 30.8 & 37 \\
\hline
\rowcolor{rowblue}\textit{\textbf{Fine-tuned on \trainset}} & & & & & & & & & & \\
Spatial-MLLM SFT\fr{16} & 60.0 & 66.0 & \textbf{90.0} & 90.0 & 62.0 & 74.0 & \textbf{64.0} & \underline{56.0} & 70.3 & 3 \\
\videojev-4B SFT\fr{20} & \underline{65.3} & \underline{72.0} & \underline{84.7} & \underline{92.7} & \underline{72.7} & \underline{75.3} & \underline{57.3} & \textbf{66.7} & \underline{73.3} & 2 \\
\textbf{\geojev (ours)}\fr{20} & \textbf{68.7} & \textbf{81.3} & 84.0 & \textbf{94.0} & \textbf{73.3} & \textbf{80.0} & 56.0 & \textbf{66.7} & \textbf{75.5} & 1 \\
\bottomrule
\end{tabular}}
\end{table*}

\paragraph{Text-only runs.}
\Cref{tab:textonly} reports results for three models given no frames. On every type, their scores are at most one point above the answer prior. All three choose the ``cannot be determined'' distractor on every Light owner item, and Qwen3-VL-32B answers ``4 or more'' on 291 of 365 Count items.

\begin{table}[h]
\centering
\caption{\textbf{Text-only runs} (accuracy \%; no frames, same prompt).}
\label{tab:textonly}
\setlength{\tabcolsep}{2.6pt}
\resizebox{\columnwidth}{!}{%
\begin{tabular}{@{}lcccccccc@{}}
\toprule
 & Count & Re-ID & Light & Patient & Device & Direction & Route & Pass. \\
\midrule
Answer prior & 34.0 & 39.5 & 46.0 & 54.8 & 39.1 & 28.3 & 30.0 & 54.3 \\
Qwen3-VL-8B text-only & 32.9 & 40.0 & 0.0 & 46.1 & 32.0 & 25.9 & 29.5 & 52.1 \\
Qwen3-VL-32B text-only & 9.9 & 38.5 & 0.0 & 52.2 & 22.7 & 25.7 & 26.1 & 49.0 \\
InternVL3.5-8B text-only & 32.9 & 29.2 & 0.0 & 55.3 & 39.1 & 26.3 & 24.7 & 53.9 \\
\bottomrule
\end{tabular}}
\end{table}

\paragraph{Proprietary models.}
For the main table, the models receive the benchmark's prompt and answer directly with reasoning off. A JSON schema constrains each reply to one option letter. We provide 10 uniformly sampled frames plus named frames as above, all at 480$\times$288. We queried \texttt{gpt-5.5} (reasoning effort \texttt{none}), \texttt{claude-sonnet-5-5} (no thinking) and \texttt{gemini-3.6-flash} (thinking budget 0) on 2026-11-15/16. One unparsable Gemini reply is scored wrong.

\begin{figure*}[!tp]
\centering
\includegraphics[width=\textwidth]{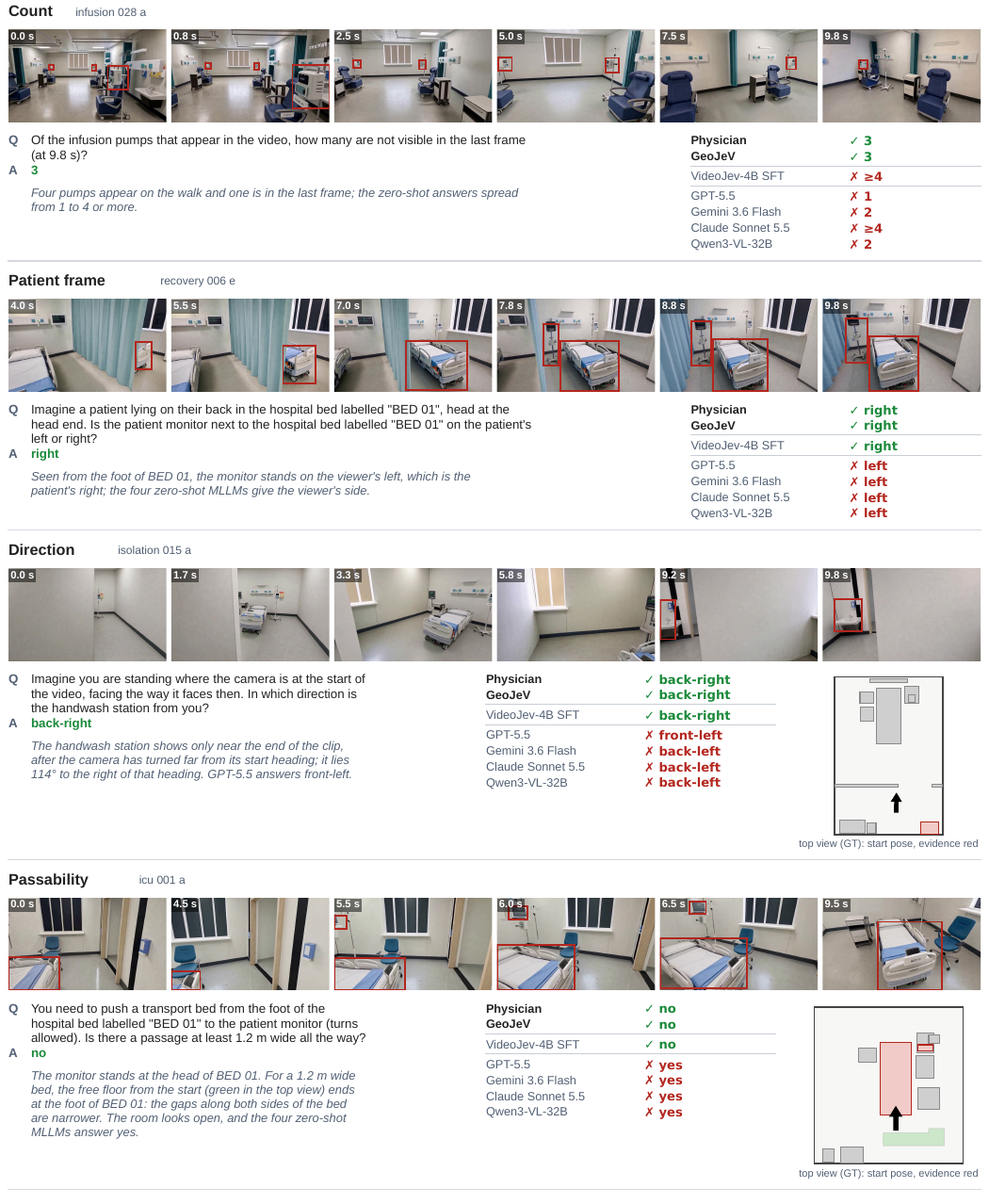}
\caption{\textbf{Case studies} on the physician subset. Each row shows frames with the ground-truth evidence boxed in red, the question (Q), its answer (A) and a note on the failure. Each answerer's choice is green if correct and red if wrong. The
Direction and Passability rows add a ground-truth top view with the start pose; green marks the floor the transport bed can reach.}
\label{fig:cases}
\end{figure*}

\paragraph{Case studies.}
\Cref{fig:cases} shows four questions from the physician subset. In each case, the physician and \geojev answer correctly, but four zero-shot MLLMs in the main table do not: GPT-5.5, Gemini 3.6 Flash, Claude Sonnet 5.5 and Qwen3-VL-32B. The figure shows their answers and that of \videojev-4B SFT. Model answers follow the main-table protocol, with one seed for each fine-tuned model. These cases illustrate failures, not how often each occurs.

\paragraph{Training data.}
With one seed and the same recipe, fine-tuning \videojev-4B on 10, 25, 50 and 100\% of the training layouts gives macro accuracies of 51.1, 63.7, 67.7 and 73.3, respectively. The main table reports the mean of three seeds, 73.9.

\paragraph{Source renders.}
On the source renders before photorealistic transfer, on which they were trained, the fine-tuned models score higher. Spatial-MLLM SFT gains 1.7 points to reach 71.6, \videojev-4B SFT gains 3.3 points to reach 77.2, and \geojev gains 3.9 points to reach 79.6.
On renders \geojev is 2.4 points above \videojev-4B SFT (paired 95\% CI [+1.3, +3.6]).

\paragraph{Paired states.}
\Cref{tab:paired} reports results for the 204 groups in which the correct answer changes between states. The unchanged-answer rate is the share of groups in which the model chooses the same option in every state and is therefore wrong at least once. Uniform random choice gives an unchanged-answer rate of 14.0\%. Group consistency is the share of groups answered correctly in every state. As a model may also repeat an answer it simply prefers, we compare the rate with that expected from its answer preferences alone (Habit in \cref{tab:paired}). Over the 23 open zero-shot models evaluated when we ran this
analysis, the excess is 3.1--41.1 points (median 14.8; all 32: same range, median 13.3), against $-$0.5 to +4.3 for the
fine-tuned \videojev-4B (three seeds).

\begin{table}[t]
\centering
\caption{\textbf{Paired states} (204 answer-changing groups). Unchanged: the same answer (option text) chosen in every state.
Habit: the same, with each state's question replaced by a random question of the same subtask and option count from
another layout (200 draws), i.e.\ what the model's answer preferences alone give. Replacements have the same option texts in all but 15 of 197,869 cases (Light owner); requiring this changes no excess by over 0.7. Excess = unchanged $-$ habit, computed before rounding. All correct:
every state answered correctly (chance 4.1\%). Proprietary models, then seven of the 23 open zero-shot models in the paired analysis (incl.\ the
lowest and highest excess), zero-shot groups by accuracy, then fine-tuned.}
\label{tab:paired}
\footnotesize
\setlength{\tabcolsep}{3pt}
\resizebox{\columnwidth}{!}{%
\begin{tabular}{@{}lcccc@{}}
\toprule
Model & Unchanged & Habit & Excess & All correct \\
\midrule
Gemini 3.6 Flash & 37.7 & 28.5 & +9.3 & 23.0 \\
GPT-5.5 & 28.9 & 24.8 & +4.1 & 22.5 \\
Claude Sonnet 5.5 & 29.4 & 22.1 & +7.3 & 15.7 \\
\midrule
\videojev-4B & 45.1 & 36.3 & +8.8 & 16.7 \\
Qwen3-VL-32B & 42.2 & 26.8 & +15.4 & 15.2 \\
InternVL3-78B & 40.7 & 30.8 & +9.9 & 17.2 \\
Qwen3-VL-8B & 49.5 & 30.8 & +18.7 & 18.6 \\
Spatial-MLLM-135K & 70.1 & 45.3 & +24.7 & 2.0 \\
MiMo-VL-7B-RL & 31.4 & 28.2 & +3.1 & 7.4 \\
SpaceLLaVA-13B & 95.1 & 54.0 & +41.1 & 1.0 \\
\midrule
Spatial-MLLM SFT & 28.4 & 20.0 & +8.4 & 37.7 \\
\videojev-4B SFT (3 seeds) & 21.1 & 19.6 & +1.5 & 45.4 \\
\geojev (3 seeds) & 21.8 & 19.3 & +2.5 & 48.9 \\
\bottomrule
\end{tabular}}
\end{table}

\section{\geojev details}
\label{sec:supp_geojev}

\paragraph{3D-input variants.}
The four 3D-input models in the main paper's shuffle-twin table use the \videojev-4B SFT recipe with one epoch and three seeds. The variants with bird's-eye-view (BEV) tokens and local BEV reads use the scene memory's frozen Lift-Splat encoder. \emph{BEV tokens}: the encoder's 256 BEV feature tokens join the visual tokens through a zero-initialised projector. \emph{Local BEV reads}: the MLLM points to up to three cells of the 32$\times$32 map, supervised by the cells of the question's evidence objects, and receives each cell's map features, position and decoded objects as tokens. \emph{Geometry head}: a bilinear head scores each option from the MLLM's hidden states and the VGGT-lifted 3D positions of detected objects; its score is added to the option's logit, and it is trained jointly with its own cross-entropy loss. \emph{GT BEV at GT positions}: local reads on ground-truth semantic maps of the objects seen in the clip. At test time only, each model's shuffle twin receives the 3D input of a clip of another layout, the one half the sorted clip list away. This donor rule differs from the same-family, same-state donor of \geojev's own shuffle control in the main paper.

\paragraph{Scene memory.}
\begin{itemize}[leftmargin=*,itemsep=1pt,topsep=2pt]
\item \emph{Depth and pose.} VGGT-1B~\cite{vggt} runs on all 60 frames at
  518$\times$294, while the MLLM's input is the 20 frames marked (20\,f) in the main table.
  When we rebuild the scene memory from the 20 base frames (every third frame, without the frames added for Re-ID) and refit the encoders and map reader, \geojev scores 75.4, compared with 75.5 for the same rebuild using all 60 frames. Across three seeds, the difference is $-0.1$ (95\% CI $[-1.2,+0.9]$). The main table reports 75.7 for the original build and 73.9 for \videojev-4B SFT. VGGT gives neither metric scale nor gravity, so we estimate both. We use RANSAC to find a horizontal plane in the lower 40\% of each frame's points, with its normal within 30\textdegree{} of the first camera's up direction. We take this normal as up and the lowest dense layer of points below the first camera as the floor. We scale the scene so that the first camera is 1.55\,m above the floor, the walkthroughs' eye height. The frame has its origin on the floor under the first camera, $y$ along its view direction, $x$ to its right and
  $z$ up.
\item \emph{Objects and tracks.} A YOLO detector~\cite{yolo} trained on training-split renders gives 2D boxes at confidence $\geq$0.3. We lift each box to 3D using its front-surface depth and the camera pose. We build tracks once per clip by greedily linking detections of the same class across frames in 3D. Links must be under 1.0\,m, and each track can have at most one detection per frame. A track counts if it spans $\geq$2 frames or has a detection at the query time.
\item \emph{Semantic BEV map.} A Lift-Splat encoder~\cite{liftsplat} uses VGGT geometry to lift the frozen \videojev visual tokens into a grid of 64$\times$64 cells of 0.25\,m with 4 height bins. It predicts class centres with a CenterPoint head~\cite{centerpoint}. We pretrain the encoder on training-split 3D boxes. The decoded 3D boxes represent the objects in the scene memory. \texttt{free\_space} uses their footprints as obstacles unless the bottom of the box is above 1.9\,m. The map $\hat M$ read by \texttt{geometry} has 32$\times$32 cells of 0.5\,m with 47 class-centre heat
  maps (2$\times$2 max-pooled) and three occupancy channels (outside / free / occupied). We use 4-fold cross-fitting by layout for the training-clip maps. Each map therefore comes from an encoder trained without that layout.
\end{itemize}

\begin{table}[t]
\centering
\caption{\textbf{\geojev's primitives.} The arguments of \texttt{memory} and \texttt{free\_space} come from one shared set of argument heads on the MLLM.}
\label{tab:primitives}
\footnotesize
\setlength{\tabcolsep}{3pt}
\resizebox{\columnwidth}{!}{%
\begin{tabular}{@{}lll@{}}
\toprule
Primitive & Arguments & Returns \\
\midrule
\texttt{memory} & class, scope, time $t$ [, $t'$, side] & count out of view / in all; identity across times \\
\texttt{free\_space} & start, goal [, facing object, width $w$] & bottleneck width; turns of the shortest route \\
\texttt{geometry} & none (question and option text) & map reader's option scores from BEV reads \\
\bottomrule
\end{tabular}}
\end{table}

\paragraph{\texttt{memory}.}
\Cref{tab:primitives} summarises the primitives. The count scope (out of view or in all) and the reference side (left or right) are arguments. Out of view at $t$: $\hat n$ = tracks seen up to $t$ minus tracks detected at $t$, clipped to $[0, 4]$; option $k$
scores $-|k - \hat n|$. In all: the number of tracks, clipped to the largest option. For identity, the reference is the detection of the specified class on the given side at $t$. Candidate $j$ at $t'$ scores $\log\softmax_j(-d_j/\tau)$, where $d_j$ is its 3D distance to the reference and $\tau = 0.3$\,m. The option ``not in that frame'' enters the softmax at a distance of 1.0\,m. An option whose ordinal exceeds the number of detections scores $-\infty$.

\paragraph{\texttt{free\_space}.}
We represent the walkable floor on a 0.1\,m grid over the floor extent of the BEV map and block cells outside that extent. We treat object boxes as obstacles and inflate them by 0.3\,m, half the walker's width. Start and goal objects are given as classes and located only if the scene memory holds one object of the class; otherwise (\eg a labelled bed in a multi-bed room) \texttt{free\_space} abstains and \geojev answers with the MLLM alone. The start is the camera's first position or, like the goal, the reachable cell nearest the object's uninflated footprint (for a bed's foot, nearest the bed), at most 0.6\,m away.
\emph{Route}. When the options are turn sequences, we run A* on the grid and simplify the path with Ramer--Douglas--Peucker at a tolerance of 0.25\,m. We read the path's first move relative to the start heading (no turn under 45\textdegree, turning around over 135\textdegree). We then take the sharpest bends in walking order so that the sequence has as many moves as the options. We choose the option that matches the most moves, breaking ties in favour of the first tied option. It scores 0 and all others score $-\infty$ in Eq.~(1) of the main paper. When we find a route, its answer therefore replaces the MLLM's. \emph{Width}. The bottleneck width $\hat w$ is the largest body width in [0.3, 2.6]\,m at which the start and goal remain connected. We find it by bisection with obstacle inflation $\hat w/2$. At width $w$, ``yes'' scores $(\hat w - w)/(2\sigma)$ and ``no'' scores its negative, with $\sigma = 0.15$\,m; the scores are added to the MLLM's log-probabilities. Width answers the 459 passage-width questions; the router sends all 57 straight-route questions to \texttt{geometry}, which checks neither the marked line nor the top clearance. We choose the reach of 0.6\,m on the validation split. On that split, accuracy is unchanged when we vary the walker's half width (0.25--0.35\,m) and turn threshold (30--60\textdegree).

\paragraph{\texttt{geometry}: map reader.}
A 2-layer transformer encodes the question and options from frozen token embeddings. The head
then makes $K = 3$ sequential hard reads of a map $M$ ($\hat M$ for the student, $M^\star$ for the teacher), each a
categorical choice over its 1,024 cells,
\begin{multline}
\rho(c_k \mid c_{<k}, q, M) \propto \\ \exp\!\Big(\tfrac{u_k(c_{<k}, q, M)^\top f_M(c_k)}{\sqrt{d}} + \log\big(H^M_{n_k(q)}(c_k) + 0.01\big)\Big),
\label{eq:read}
\end{multline}
where $u_k$ is a query conditioned on the features and positions of earlier reads, $f_M$ embeds map cells, and $d$ is the feature dimension. The second term, the name prior, uses the heat map of the $k$-th class $n_k(q)$ named in the question. We omit this term if the question names fewer than $k$ classes. Reads are
sampled in training and taken by argmax at test time. Each option attends softly over all 1,024 cells and a learned null slot. An MLP scores each option using the question and option text, the option's own soft read, the $K$ hard reads, and the offsets, distances and bearings between the reads. The log-softmax over options gives $s(o)$.

\paragraph{Privileged teacher and on-policy distillation (OPD).}
A teacher with the same architecture reads the ground-truth
map $M^\star$, whose class channels mark full object footprints rather than predicted centres. We train the teacher first, forcing its reads to the cells of the objects that determine the answer. The loss includes read cross-entropy with weight 0.5, answer cross-entropy and the map-dependence term below. We then freeze the teacher. Training the student with forced reads would condition each read on correct earlier reads. The student rarely makes those correct reads on a noisy predicted map. Following DAgger~\cite{dagger}, we instead sample read prefixes from the student. We use a privileged teacher as in Learning by Cheating~\cite{lbc} and match the student's next-read distribution to the teacher's at every visited prefix:
\begin{multline}
\mathcal L = \mathrm{CE}(p_S, y) \\
+ \mathbb E_{c \sim \sg(\rho_S)} \tfrac1K \sum_{k} \mathrm{KL}\big(\rho_T(\cdot \mid c_{<k}, M^\star) \,\|\, \rho_S(\cdot \mid c_{<k}, \hat M)\big) \\
+ \mathrm{KL}(p_T \,\|\, p_S) + \mathrm{KL}\big(\mathcal U \,\|\, p_S(\cdot \mid \hat M_{\mathrm{shuf}})\big),
\label{eq:opd}
\end{multline}
where $p_S$ and $p_T$ are the student's and the teacher's answer distributions (the teacher with its own greedy reads), $y$ is the answer, $\mathcal U$ is uniform over the options and $\sg$ stops gradients through sampling. The last term pushes the student toward a uniform answer when given another clip's map $\hat M_{\mathrm{shuf}}$, discouraging answers from text alone. All weights in \cref{eq:opd} are 1. The head trains for 30 epochs (AdamW, learning rate $10^{-3}$, one-cycle schedule, batch 64).

\paragraph{Router and argument heads.}
The MLLM's question state $h_q$ is the query input to \videojev's answer-scoring decision head. We extract it from the frozen fine-tuned model for the 28,466 training questions that at least one primitive answers correctly on its own. These correct primitives, without \texttt{geometry} when \texttt{memory} or \texttt{free\_space} is among them, form the router's target set $\mathcal S$. We train the router with the marginal likelihood $-\log \sum_{\pi \in \mathcal S} p(\pi \mid h_q)$. The training questions supply the argument targets, but at test time the heads see only $h_q$. Each head is a
single linear layer trained with AdamW (learning rate $10^{-3}$, batch 512). For each seed, we choose the weight decay and number of epochs using a held-out 10\% of the training clips. We then refit the heads on all training clips. The argument heads for \texttt{memory} and \texttt{free\_space} predict the counted object, start, facing object and goal, each with 47 classes. The start may also be the camera; the facing object applies to an object start. The heads also predict the side of the reference, the count scope (out of view or in all), and which numbers in the question give the time, second time and width.

\section{Limitations}
\label{supp:limitations}

All scenes are synthetic, static and unoccupied. The photorealistic transfer depends on hosted generators and is intended to change appearance, not content. We leave a real-video subset to future work. Training and test layouts share the generator, assets and templates. The fine-tuned results are therefore in-distribution for scene layout and question form, but not for appearance, since the training clips are renders. \geojev's argument heads are trained on
template-derived targets, and its primitives rely on estimated geometry.